\documentclass{article}

\usepackage{microtype}
\usepackage{graphicx}
\usepackage{subcaption}
\usepackage{booktabs} 

\usepackage{hyperref}

\usepackage[accepted]{icml2026}

\usepackage{amsmath}
\usepackage{amssymb}
\usepackage{mathtools}
\usepackage{amsthm}
\usepackage{algorithm}
\usepackage{algorithmic}

\usepackage[capitalize,noabbrev]{cleveref}

\theoremstyle{plain}

\theoremstyle{definition}

\theoremstyle{remark}

\usepackage[textsize=tiny]{todonotes}

\icmltitlerunning{Counterfactual Transport Flows for Offline Conservative Trajectory Refinement}

\begin{document}

\twocolumn[
  \icmltitle{Counterfactual Transport Flows for Offline Conservative Trajectory Refinement}



  \icmlsetsymbol{equal}{*}

  \begin{icmlauthorlist}
    \icmlauthor{Lena Krieger}{equal,yyy,comp}
    \icmlauthor{Xuan Zhao}{equal,yyy}
    \icmlauthor{Zhuo Cao}{yyy}
    \icmlauthor{Qin Wang}{yyy}
    \icmlauthor{Hanno Scharr}{yyy}
    \icmlauthor{Ira Assent}{yyy,sch}
  
  \end{icmlauthorlist}

  \icmlaffiliation{yyy}{IAS-8, Forschungszentrum Jülich, Germany}
  \icmlaffiliation{comp}{LMU Munich, Munich Center for Machine Learning (MCML), Germany}
  \icmlaffiliation{sch}{Department of Computer Science, Aarhus University, Denmark}
\icmlcorrespondingauthor{Lena Krieger}{l.krieger@fz-juelich.de}
  \icmlcorrespondingauthor{Xuan Zhao}{x.zhao@fz-juelich.de}

  \icmlkeywords{Machine Learning, ICML}

  \vskip 0.3in
]



\printAffiliationsAndNotice{}  

\begin{abstract}
Offline reinforcement learning (RL) offers a path to policy improvement from logged data alone, using historical returns or other measurable outcomes as world feedback. A key difficulty is improving observed behavior without extrapolating beyond what the offline data supports. 
We propose \emph{counterfactual transport flows}, a source-conditioned trajectory refinement framework for offline decision-making guided by world feedback. Given a low-feedback candidate trajectory, 
we construct local preference pairs from offline data by retrieving nearby trajectories in latent trajectory space with higher task-specific feedback, and use them as weak supervision for conservative refinement. The framework learns instance-specific refinement directions: 
at inference time, a refinement strength parameter controls how far the candidate trajectory is transported, enabling a trade-off between preserving the original behavior and applying stronger improvement. Experiments on D4RL benchmarks, including AntMaze and MuJoCo tasks, show that our method improves behavior from historical returns as world feedback, while providing interpretable trajectory-level refinement paths.
\end{abstract}

\section{Introduction}

\begin{figure}
    \centering
    \includegraphics[clip, trim=0.85cm 3.0cm 7.05cm 2.4cm,width=0.95\linewidth]{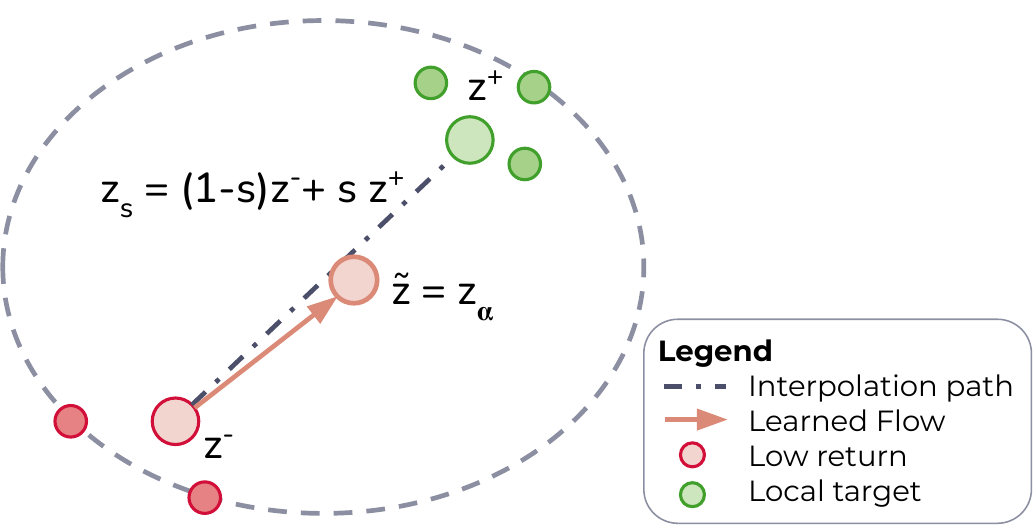}
    \caption{\textbf{Source-conditioned trajectory refinement in latent trajectory space.} 
  Given a low-feedback source $z^-$, a locally similar higher-feedback target $z^+$ is retrieved from the latent neighborhood. The dashed line shows the training interpolation
  path $z_s = (1-s)z^- + sz^+$; the orange arrow shows the learned
  source-conditioned flow. At inference,
  $\tilde{z} = z_\alpha$ is the partially refined trajectory, with
  $\alpha \in [0,1]$ controlling refinement strength.
}
    \label{fig:enter-label}
\end{figure}

Offline reinforcement learning \citep{levine2020offline} seeks to improve behavior using only logged trajectories and their observed outcomes.
Existing offline methods either constrain the policy via value penalties \citep{kumar2020conservative, kostrikov2021offline, fujimoto2021minimalist} or use diffusion and flow-based models to generate new behavior \citep{wang2022diffusion, park2025flow, feng2025offline,janner2022planning}. A separate line of work augments offline datasets with counterfactual transitions \citep{pitis2022mocoda, zhang2023flow, sun2024acamda, lu2020sample}, but operates primarily at the transition level rather than over full trajectories. These approaches improve or regularize behavior, but they do not explicitly model how a specific low-feedback trajectory could be locally revised toward a better alternative.

A common limitation of standard RL is that learned policies are optimized as reactive mappings from states to actions, compressing past experience into fast response rules. While effective for control, such policies do not explicitly reveal how an observed low-outcome behavior could have been modified to succeed. Recent critiques of world models argue that purposeful agents should support hypothetical reasoning over actionable possibilities, rather than merely reactive responses \citep{xing2025critiques, lecun2022path}. Unlike existing offline methods that learn global policies or trajectory distributions, our method learns local, source-conditioned refinement flows for individual trajectory improvement. Motivated by this perspective, we formulate offline improvement as source-conditioned trajectory refinement: given a low-feedback trajectory, our goal is to locally revise it toward a nearby alternative with improved world feedback, e.g., measurable scalar outcomes such as return, safety, or task success. Unlike reward-conditioned generation, which conditions on a desired outcome level, our method conditions on the source trajectory itself, yielding instance-specific refinement guidance.

We propose \emph{counterfactual transport flows}, which use conditional flow matching \citep{DBLP:conf/iclr/LipmanCBNL23,DBLP:journals/tmlr/0001FMHZRWB24,DBLP:journals/corr/abs-2510-14623,albergo2023building} to learn source-conditioned refinement directions in a latent trajectory space. Given a low-feedback source trajectory, we retrieve nearby trajectories in latent trajectory space with higher feedback and use them as weak local improvement targets. The learned flow transports the source trajectory toward nearby higher-feedback alternatives. A refinement strength parameter $\alpha$ controls the trade-off between conservatism and improvement. Our method is trained entirely from world feedback, requiring no human preference labels or separately trained reward models, grounding improvement in the measured consequences of past behavior. Beyond return improvement, the transport perspective provides trajectory-level interpretability. Because the learned vector field traces explicit refinement paths from source trajectories toward locally improved alternatives, it yields counterfactual-style explanations of how low-feedback behaviors may be revised.

Our contributions are: (i) we formulate offline decision improvement as source-conditioned trajectory refinement from world feedback, rather than global policy optimization or unconstrained trajectory generation; (ii) we construct local preference pairs from observed outcomes and use conditional flow matching to learn instance-specific refinement directions in latent trajectory space; and (iii) we validate the approach on D4RL \citep{fu2020d4rl} trajectories including AntMaze and MuJoCo \citep{todorov2012mujoco}, showing improved feedback with conservative trajectory deviations.

\section{Method}

Many real-world decision-making problems are not solved by generating decisions from scratch, but by revising an existing proposal using feedback from observed outcomes. A treatment plan proposed by a clinician \citep{gottesman2019guidelines}, a portfolio allocation designed by an analyst \citep{bai2025review}, or a recommendation sequence produced by a deployed system \citep{afsar2022reinforcement} may serve as an initial candidate whose consequences can be evaluated through task-specific world feedback, such as health outcomes, financial return, safety, or user engagement. In such settings, the goal is not to replace the original decision entirely, but to refine it while preserving its overall structure.

We formulate decision improvement as an instance-level trajectory refinement problem. Given an initial candidate trajectory and its observed or predicted outcome, our goal is to learn how this particular trajectory can be locally revised toward a nearby alternative with improved world feedback. Unlike global policy optimization, which learns a policy over all states, our formulation focuses on source-conditioned refinement:
$
\tau^- \longmapsto \tau^*,
$
where $\tau^-$ is the original candidate trajectory and $\tau^*$ is its locally refined version.

\subsection{Problem Formulation}

We consider a sequential decision setting where a trajectory
$
\tau = (s_0,a_0,\dots,s_T)
$
represents a sequence of states and actions. Given a candidate trajectory $\tau^-$, our goal is to obtain a refined trajectory $\tau^*$ that improves the observed world feedback while remaining close to the original candidate.

Conceptually, the desired counterfactual refinement can be written as
\begin{equation}
\label{eq:ideal_cf}
\tau^*
=
\arg\min_{\tau' \in \mathcal T_{\mathrm{feas}}}
d_\tau(\tau',\tau^-)
\quad
\text{s.t.}
\quad
R(\tau') \ge R(\tau^-)+\delta,
\end{equation}
where $\mathcal T_{\mathrm{feas}}$ denotes trajectories that remain behaviorally plausible under the offline data distribution, $d_\tau(\cdot,\cdot)$ measures trajectory-level deviation, $R(\tau)$ denotes world feedback, such as return, safety, or another measurable outcome, and $\delta \ge 0$ specifies a desired improvement margin.

Eq.~\eqref{eq:ideal_cf} captures an idealized notion of minimal-change improvement: among feasible trajectories that improve the world feedback, we seek the one closest to the original. In offline settings, however, this optimization is intractable, evaluating arbitrary trajectories requires either environment access or a reliable dynamics model, neither of which is available.
We therefore approximate it by learning a local refinement operator from offline data:
$
\tau^* = T_\theta(\tau^-).
$
The operator amortizes local improvement patterns observed in the offline data into a single forward pass, conditioned on the input trajectory $\tau^-$.

\subsection{Local Preference Pairs}

To approximate the ideal objective in Eq.~\eqref{eq:ideal_cf}, our method constructs weak local preference pairs from offline trajectories. 
For a given source trajectory $\tau^-$, we search for similar trajectories in latent trajectory space with better observed world feedback. These pairs provide examples of nearby alternatives that are preferable to the source trajectory. Our formulation is related to recent preference-based flow learning approaches \citep{DBLP:conf/nips/KimLKOCY24}, but focuses on local source-conditioned trajectory refinement in offline decision-making. Formally, let $\mathcal D$ denote the offline dataset. For each trajectory $\tau^-$, we define a local latent neighborhood
$
\mathcal N(\tau^-)
=
\{
\tau \in \mathcal D :
d_\tau(\tau,\tau^-)\le \epsilon
\},
$
where $d_\tau$ measures trajectory similarity in latent trajectory space and $\epsilon$ controls locality. We construct local preference pairs
$
(\tau^-,\tau^+),
\quad
\tau^+ \in \mathcal N(\tau^-),
$
such that
$
R(\tau^+) \ge R(\tau^-)+\delta.
$
When multiple such trajectories exist, we select a nearby higher-feedback target satisfying the improvement criterion, 
$
\tau^+
=
\arg\min_{\tau \in \mathcal N(\tau^-)}
d_\tau(\tau,\tau^-)
\quad
\text{s.t.}
\quad
R(\tau)\ge R(\tau^-)+\delta,
$
where $\delta$ is the same margin as in Eq.~\eqref{eq:ideal_cf}.

This construction approximates Eq.~\eqref{eq:ideal_cf} empirically: rather than searching over all feasible trajectories, we restrict candidates to nearby trajectories in latent trajectory space that are actually observed in the data.
The pairs are not true counterfactual optima, but provide weak supervision for local refinement, meaningful under the assumption that differences in world feedback within a neighborhood are primarily attributable to differences in trajectory-level decisions rather than unobserved variation. 
We use the term \emph{counterfactual} in the trajectory-refinement sense: $\tau^+$ represents what the agent could have done differently to achieve better feedback under the same conditions, grounded in observed data rather than a structural causal model. 

\subsection{Source-Conditioned Flow Matching}

We encode trajectories into a latent trajectory space using a trajectory encoder $\phi$, with
$
z^-=\phi(\tau^-)
$
 and 
$
z^+=\phi(\tau^+)
$, and learn a decoder $\psi$ mapping latent representations back to trajectory space.
Rather than training a one-shot predictor from $z^-$ to $z^+$, which would produce a single endpoint without exposing intermediate refinement structure or allowing control over refinement strength, we model refinement as a continuous transport process.
For each local preference pair, we define the linear interpolation
$
z_s = (1-s)z^- + sz^+,
\quad
s\sim \mathcal U(0,1),
$
with target velocity
$
u_s = z^+ - z^-.
$

Crucially, the vector field is conditioned on the source trajectory $z^-$:
$
v_\theta(z_s,s \mid z^-).
$
This anchors the transport to the particular trajectory being refined, so the model learns how this specific source can be locally revised rather than learning a global direction toward high-feedback regions. 
We train the source-conditioned vector field using the flow matching objective \citep{DBLP:conf/iclr/LipmanCBNL23,DBLP:journals/tmlr/0001FMHZRWB24,DBLP:journals/corr/abs-2510-14623,albergo2023building}:
\begin{equation}
\label{eq:source_fm}
\mathcal L_{\mathrm{FM}}
=
\mathbb E_{z^-,z^+,s}
\left[
\left\|
v_\theta(z_s,s \mid z^-)
-
(z^+-z^-)
\right\|_2^2
\right].
\end{equation}
The learned vector field represents an instance-conditioned refinement direction induced by local preference pairs, not a reward gradient or global optimization target.

\subsection{Inference with Refinement Strength}

At inference time, we assume access to an initial candidate trajectory $\tau^-$, which may be produced by a base policy, a planning module, a heuristic controller, a sequence model, or a human decision maker. 
We encode it as
$
z^-=\phi(\tau^-)
$
and integrate the learned source-conditioned vector field from $z_0=z^-$:
$
\frac{dz_s}{ds}
=
v_\theta(z_s,s \mid z^-),
\quad
s\in[0,1].
$

At test time, no improved target trajectory is required. The learned vector field amortizes the local refinement directions observed in the training preference pairs. We stop integration at a refinement strength $\alpha\in[0,1]$ yielding the partially transported latent state
$
\tilde z = z_\alpha,
$
which is decoded as
$
\tau^*=\psi(\tilde z).
$
The parameter $\alpha$ controls the degree of revision:  $\alpha=0$ recovers the original trajectory, while larger values of $\alpha$ apply stronger refinement toward the locally improved direction, enabling a continuous trade-off between conservatism and improvement.



Unlike reward-conditioned generation, where the condition specifies a desired outcome level, our method conditions on the source trajectory itself. The guidance is therefore instance-wise: it describes how a particular trajectory can be locally revised, rather than how to generate a generally high-outcome trajectory from scratch.

\begin{table*}[t]
\centering
\caption{\textbf{Feedback improvement and deviation comparison on D4RL trajectories.}  Feedback $\Delta$ measures improvement over the source trajectory, action and latent deviation measure conservatism. Source-conditioned local refinement (ours) achieves the best improvement-deviation trade-off, outperforming retrieval baselines on conservatism and non-local flow matching on both metrics. } 
\label{tab:main_results}
\resizebox{0.98\linewidth}{!}{
\begin{tabular}{lccc|ccc}
\toprule
& \multicolumn{3}{c|}{AntMaze} & \multicolumn{3}{c}{MuJoCo} \\
\cmidrule(lr){2-4} \cmidrule(lr){5-7}
Method 
& Feedback $\Delta \uparrow$ 
& Action Dev. $\downarrow$ 
& Latent Dev. $\downarrow$
& Feedback $\Delta \uparrow$ 
& Action Dev. $\downarrow$ 
& Latent Dev. $\downarrow$ \\
\midrule

Nearest improved Neighbor 
& +62.50 & 1.83 & 38.70
& +1050.30 & 2.14 & 42.50 \\

Random improved Trajectory 
& +117.50 & 2.18 & 62.50
& +1820.40 & 2.56 & 68.20 \\

Non-local Flow Matching 
& -11.48 & 1.89 & 57.30
& -210.70 & 1.92 & 52.10 \\

\midrule

Ours ($k=3$)
& +69.44 & 1.41 & 27.00
& +1180.20 & 1.58 & 31.40 \\

\bottomrule
\end{tabular}
}
\end{table*}

\section{Experiments}
We evaluate transport flows on standard offline benchmarks, assessing trajectory refinement quality and learned-flow conservatism rather than policy performance.
\paragraph{Datasets and Setup.}
We evaluate our method on offline trajectory datasets from D4RL \citep{fu2020d4rl}, including AntMaze and HalfCheetah tasks from the MuJoCo \citep{todorov2012mujoco} suite. These datasets contain trajectories of varying quality collected by heterogeneous behavior policies, making them suitable for studying local trajectory refinement. We train a trajectory autoencoder and use the learned latent space for local pair construction, flow matching, and trajectory refinement. For each source trajectory, we construct a local latent neighborhood using top-$k$ nearest-neighbor retrieval and select nearby trajectories satisfying the improvement-margin criterion as local refinement targets. 
Unless otherwise specified, all main experiments use $k=3$, which empirically provides the best trade-off between refinement quality and conservatism (Appendix Table~\ref{tab:k_ablation}). At inference time, we refine the source trajectory using the learned source-conditioned flow. Unless otherwise specified, we use refinement strength $\alpha=1.0$ and analyze the effect of varying $\alpha$ in Appendix Table~\ref{tab:alpha_ablation}. 
Since our focus is offline trajectory refinement rather than closed-loop policy deployment, evaluation is performed 
 on refined trajectories using feedback improvement and deviation metrics, without environment interaction.

\paragraph{Evaluation Metrics.}
We evaluate both improvement and conservatism. Improvement is measured by feedback $\Delta$, the predicted increase in trajectory-level feedback from a separately trained return predictor on held-out offline trajectories. Conservatism is measured by action and latent deviation between refined and source trajectories. Together, these metrics assess whether refinement improves predicted feedback while remaining close to the offline behavior distribution. 

\paragraph{Baselines.}
We compare against three baselines that isolate the contributions of locality, source conditioning, and continuous refinement. \emph{Nearest improved neighbor} replaces the source trajectory with the nearest latent-space trajectory satisfying the improvement criterion. \emph{Random improved trajectory} replaces the source with a randomly selected higher-feedback trajectory from the offline dataset, isolating the effect of local nearest-neighbor target selection. \emph{Non-local Flow Matching} trains the same source-conditioned flow model but removes locality constraints during target construction, allowing us to assess whether local target construction is necessary for conservative refinement.





\paragraph{Results and Analysis.}
\cref{tab:main_results} reports feedback improvement and trajectory deviation on AntMaze and MuJoCo. Across both domains, our method achieves the most favorable improvement--deviation trade-off. On AntMaze, source-conditioned local refinement improves feedback by $+69.44$ while producing the lowest action ($1.41$) and latent deviation ($27$) of any method. Direct neighbor-replacement baselines achieve larger raw feedback gains but induce substantially larger deviations from the source trajectories: random improved trajectory obtain $+117.50$ feedback improvement but increase action and latent deviation to $2.18$ and $62.50$, respectively. This suggests that directly replacing trajectories with improved neighbors does not yield conservative refinement. Non-local flow matching degrades feedback ($-11.48$) despite large deviations, indicating that locality is necessary for learning stable refinement directions rather than globally averaged transport dynamics.

On MuJoCo, the pattern is consistent. Our method achieves $+1180.20$ feedback improvement with action and latent deviation of $1.58$ and $31.40$, outperforming the nearest improved neighbor ($+1050.30$ feedback, $2.14/42.50$ deviation) on the improvement--deviation trade-off despite the neighbor achieving lower raw improvement. 
Random improved trajectory again achieves the highest raw gain ($+1820.40$) but induces the largest deviations ($2.56/68.20$), and non-local flow matching again degrades performance ($-210.70$). Together, these results suggest that local target construction and source conditioning are both necessary for conservative trajectory refinement. 
Additional ablations in Appendix Tables~\ref{tab:k_ablation} and~\ref{tab:alpha_ablation} further show that moderate local neighborhood sizes provide the best refinement stability, while increasing refinement strength $\alpha$ yields a controllable trade-off between improvement and conservatism.

\section{Discussion and Conclusion}

We introduced counterfactual transport flows, a source-conditioned trajectory refinement framework for offline decision-making from world feedback. Rather than generating high-outcome trajectories from scratch, 
our method locally revises an existing candidate using preference pairs constructed from scalar returns, learned via source-conditioned flow matching in a latent trajectory space. The refinement strength parameter $\alpha$ enables conservative-to-aggressive editing, and the framework operates as a plug-in operator where a candidate trajectory and world feedback are available. 

On D4RL benchmarks including AntMaze and MuJoCo, source-conditioned refinement improves offline trajectory feedback relative to the source trajectories using only scalar returns, no human preference labels, reward models, or online interaction required.
More broadly, our results suggest that trajectory-level refinement from world feedback is a promising direction for reliable offline improvement. 
Several directions for future work remain open. 
First, local preference construction currently relies on $k$-nearest neighbor retrieval, learned or adaptive neighborhood selection could improve robustness, particularly as refinement quality depends on whether the learned representation preserves meaningful local behavioral structure. 
Second, feasibility guarantees for decoded trajectories in latent transport and theoretical conditions under which source-conditioned refinement leads to consistent outcome improvement remain important open questions.
Finally, integrating the framework with learned world models would enable iterative hypothetical revision and counterfactual planning over candidate futures rather than purely reactive policy execution.
In our work, we use the term ``counterfactual'' in the trajectory-refinement sense rather than in the formal causal inference sense. 
\clearpage
\section*{Acknowledgements}
This work was partially funded by project W2/W3-108 Initiative and Networking Fund of the Helmholtz Association. We gratefully acknowledge the computing time granted through project XAI (No. 65881) on the supercomputer JURECA at Jülich Supercomputing Centre (JSC).
\bibliography{iclr2026_conference}
\bibliographystyle{icml2026}

\newpage
\appendix
\onecolumn

\section{Experimental Details}

\subsection{Datasets}

We evaluate the proposed method on D4RL benchmark trajectories \citep{fu2020d4rl}, focusing on MuJoCo \citep{todorov2012mujoco} locomotion and AntMaze navigation tasks. Each dataset consists of offline trajectories collected by previously trained policies with varying behavior quality and coverage.

For MuJoCo tasks, trajectories contain continuous state-action transitions with dense return feedback. For AntMaze, trajectories correspond to long-horizon navigation behavior with sparse rewards and heterogeneous success quality. In all experiments, trajectories are segmented into fixed-length windows before encoding and refinement.

\subsection{Trajectory Representation}

Each trajectory
$
\tau=(s_0,a_0,\dots,s_T)
$
is represented as a sequence of continuous states and actions. We train a trajectory autoencoder to map trajectories into a latent trajectory space:
$
z=\phi(\tau).
$

The encoder and decoder are implemented as multilayer perceptrons operating on flattened trajectory representations. Reconstruction training uses mean squared error between original and reconstructed trajectory tokens. The latent representation is used both for local preference retrieval and for source-conditioned flow matching.

\subsection{Local Preference Pair Construction}

Local preference pairs are constructed directly from offline trajectories using observed trajectory returns. For each source trajectory $\tau^-$, we retrieve nearby latent trajectories using Euclidean distance in latent space.

Given a source latent trajectory $z_i$, we identify candidate targets whose observed return exceeds that of the source trajectory by at least a minimum improvement margin:
$
R(\tau_j)-R(\tau_i)>\delta.
$

Among these candidates, the nearest higher-feedback trajectory is selected as the local improvement target. The neighborhood size $k$ controls the locality of retrieval. Unless otherwise stated, the main experiments use highly local neighborhoods ($k=3$), which empirically provide the most stable refinement behavior.

\subsection{Source-Conditioned Flow Matching}

The source-conditioned flow model learns a vector field
$
v_\theta(z_s,s\mid z^-)
$
using conditional flow matching. For each local preference pair $(z^-,z^+)$, interpolation points are sampled as
$
z_s=(1-s)z^-+sz^+,
\quad
s\sim\mathcal U(0,1),
$
with target velocity
$
u_s=z^+-z^-.
$

The vector field is parameterized by a multilayer perceptron conditioned on the interpolated latent state $z_s$, interpolation time $s$, and source latent trajectory $z^-$. Training minimizes mean squared error between predicted and target velocities.

At inference time, refinement is performed by integrating the learned vector field starting from the source latent trajectory:
$
\frac{dz_s}{ds}=v_\theta(z_s,s\mid z^-).
$

A refinement-strength parameter $\alpha\in[0,1]$ controls the amount of transport, where $\alpha=0$ preserves the original trajectory and larger values apply stronger refinement.

\subsection{Offline Proxy Evaluation}

Because the current study focuses on offline trajectory refinement rather than closed-loop policy deployment, evaluation is performed using offline proxy metrics. Accordingly, trajectory quality is evaluated using an offline proxy-feedback predictor trained on held-out trajectories rather than environment rollouts. Future work should further investigate whether the learned refinement directions translate to improved closed-loop control performance.

We train a separate return predictor on held-out latent trajectories to estimate trajectory-level outcome feedback. Importantly, this predictor is used only for evaluation and is not used for local pair construction or flow training.

We report proxy feedback improvement, defined as the predicted improvement in trajectory-level feedback relative to the source trajectory. We additionally measure action deviation between refined and source actions to quantify conservatism in action space, as well as latent deviation between refined and source latent trajectories.

\subsection{Baselines}

We compare the proposed method against several trajectory-level baselines. The nearest improved neighbor baseline directly retrieves the nearest trajectory with higher observed return. The random improved trajectory baseline randomly selects a higher-feedback trajectory. We additionally compare against non-local flow matching, which removes locality constraints during target construction, and autoencoder reconstruction without refinement.

These baselines allow us to evaluate the trade-off between improvement and conservatism, as well as the importance of locality in trajectory refinement.

\begin{table}[t]
\centering
\caption{\textbf{Ablation on neighborhood size $k$.} $k=1$ corresponds to nearest higher-return neighbor retrieval. Our method uses $k=3$ in main experiments.}
\label{tab:k_ablation}
\begin{tabular}{lccc}
\toprule
$k$ & Feedback$\Delta$ $\uparrow$ & ActionDev $\downarrow$ & LatentDev $\downarrow$ \\
\midrule
1 (Nearest) & 62.50 & 1.830 & 38.70 \\
3           & 69.44 & 1.406 & 27.79 \\
5           & 51.39 & 1.392 & 27.55 \\
10          & 24.41 & 1.387 & 27.18 \\
20          & 8.55  & 1.395 & 28.31 \\
50          & 32.30 & 1.435 & 30.25 \\
\bottomrule
\end{tabular}
\end{table}

\begin{table}[t]
\centering
\caption{\textbf{Effect of refinement strength $\alpha$ on improvement and conservatism.} As $\alpha$ increases, feedback improvement (Feedback$\Delta$) and deviations (ActionDev, LatentDev) increase monotonically. Our main experiments use $\alpha=1.0, k=3$.}
\label{tab:alpha_ablation}
\begin{tabular}{lccc}
\toprule
$\alpha$ & Feedback$\Delta$ $\uparrow$ & ActionDev $\downarrow$ & LatentDev $\downarrow$ \\
\midrule
0.000 & 0.000 & 0.000 & 0.000 \\
0.250 & 8.137 & 1.246 & 5.768 \\
0.500 & 17.400 & 1.262 & 13.218 \\
0.750 & 27.395 & 1.331 & 22.193 \\
1.000 & 69.440 & 1.455 & 27.790 \\
1.250 & 72.150 & 1.619 & 44.203 \\
\bottomrule
\end{tabular}
\end{table}

\subsection{Implementation Details}

All models are implemented in PyTorch. Optimization uses Adam with minibatch stochastic gradient descent. Latent trajectory embeddings are normalized before flow training. During evaluation, refinement trajectories are generated by numerical integration of the learned vector field using fixed-step Euler updates.

Unless otherwise specified, experiments use latent-space nearest-neighbor retrieval with neighborhood size $k=3$ and refinement strength $\alpha=1.0$.

\end{document}